\documentclass[a4paper,conference]{IEEEtran}

\ifCLASSINFOpdf
\else
\fi

\usepackage{multirow}
\usepackage{times}

\usepackage{soul}
\usepackage{url}
\usepackage[hidelinks]{hyperref}
\usepackage[utf8]{inputenc}
\usepackage[small]{caption}
\usepackage{graphicx}
\usepackage{amsmath}
\usepackage{booktabs}
\usepackage{comment}
\usepackage{color}
\usepackage{multirow}
\usepackage{subcaption}
\newcommand*\rfrac[2]{{}^{#1}\!/_{#2}}
\begin{document}
%
\title{Temporal Collaborative Filtering with Graph Convolutional Neural Networks}

\author{\IEEEauthorblockN{Esther Rodrigo Bonet, Duc Minh Nguyen and Nikos Deligiannis}
\IEEEauthorblockA{ETRO Department, Vrije Universiteit Brussel, Pleinlaan 2, B-1050 Brussels, Belgium\\
imec, Kapeldreef 75, B-3001, Leuven, Belgium  \\
\{erodrigo, mdnguyen, ndeligia\}@etrovub.be}}


%


\maketitle

\begin{abstract}
Temporal collaborative filtering (TCF) methods aim at modelling non-static aspects behind recommender systems, such as the dynamics in users' preferences and social trends around items. State-of-the-art TCF methods employ recurrent neural networks (RNNs) to model such aspects. These methods deploy matrix-factorization-based (MF-based) approaches to learn the user and item representations. Recently, graph-neural-network-based (GNN-based) approaches have shown improved performance in providing accurate recommendations over traditional MF-based approaches in non-temporal CF settings. Motivated by this, we propose a novel TCF method that leverages GNNs to learn user and item representations, and RNNs to model their temporal dynamics. A challenge with this method lies in the increased data sparsity, which negatively impacts obtaining meaningful quality representations with GNNs. To overcome this challenge, we train a GNN model at each time step using a set of observed interactions accumulated time-wise. Comprehensive experiments on real-world data show the improved performance obtained by our method over several state-of-the-art temporal and non-temporal CF models.
\end{abstract}

%
\IEEEpeerreviewmaketitle

\section{Introduction}

Recommender systems aim to provide users with the most relevant information or products
, with which they are likely to interact. Providing such information helps users quickly navigate and filter out irrelevant information from the plethora of data available online nowadays. As such, recommender systems have become indispensable components of online platforms, such as e-commerce, movie streaming and news websites, to drive user engagement and interactions. 

Building recommender systems has been a very active research topic for years, resulting in a great number of methods proposed in the literature. Among them, collaborative filtering (CF) methods are arguably the most popular ones due to favorable performance compared to other methods~\cite{cfdos}. CF methods leverage collective user-item interactions and build models to predict the likelihood of unobserved interactions. For instance, models that follow the matrix-factorization (MF) approach --a very common CF approach-- learn users' and items' latent representation vectors (also referred to as \textit{latent embeddings or states}) by factorizing data matrices which contain historical interactions. Unknown user-item interaction scores are then calculated by taking the dot product of the corresponding representation vectors. These vectors, or embeddings, are often interpreted as encoding preferences of users and characteristics of items.

Despite being widely-adopted, traditional CF methods focus mainly on \textit{static settings}, where user preferences and other factors such as social trends around items, are \textit{assumed to be stationary}. In real application settings, however, such assumption seldom holds~\cite{rrn2019}. Recently, numerous works have focused on modeling the temporal dynamics in recommender systems~\cite{deng2018,Koren2010,song}. They are referred to as time-aware or 
temporal collaborative filtering (TCF) methods~\cite{tcf,Liu2013,timereview}. TCF methods often employ recurrent neural networks (RNNs) to model the temporal trajectories of user embeddings~\cite{song,fusion}, or of both user and item embeddings ~\cite{rrn2019,deng2018,ntf}. They have achieved higher performance over non-temporal counterparts in predicting future ratings~\cite{rrn2019,Koren2010,Liu2013}. 


A limitation of existing TCF methods is that they often rely
on linear MF models to individually learn users' and items' embeddings,  neglecting the fact that correlations amongst users and items are effective hints in modeling their latent representations. In contrast, recent studies on static
CF have shown the benefits of modeling user-item
interactions in forms of graphs and of using graph neural
networks (GNNs) to learn representations over linear models~\cite{Rianne2017,bnp}. 
Motivated by this, we aim to leverage GNNs in the context of \textit{temporal} CF. To this end, we propose a method that (\textit{i}) employs a graph-based CF model, namely the Graph Convolutional Matrix Completion (GCMC) model~\cite{Rianne2017}, to effectively learn user and item embeddings and (\textit{ii}) models the temporal trajectories of these embeddings using RNNs. The RNNs, after training, can propagate the latent embeddings to a future time step. These latent states are then used to predict potential user-item interactions. An inherent challenge with this method in the temporal CF setting lies in the increased data sparsity. To overcome the challenge of efficiently learning the embeddings from highly sparse data, we propose to use the historical user-item interactions, accumulated over time. Our experimental results show systematic performance improvement obtained with this approach. 

To summarize, our contributions in the paper are two-fold:
\begin{itemize}
    \item We propose a method, coined Time-aware Graph-based Matrix Completion (TG-MC), which leverages the state-of-the-art graph-based CF model with RNNs for collaborative filtering. To the best of our knowledge, we are the first to leverage the graph-based CF approach for TCF.
    \item We present comprehensive experiments on large-scale real-world data to assess the effectiveness of our method in comparison with the state of the art.
\end{itemize}

The remainder of the paper is organized as follows: Section II reviews the related work, Section III presents our method in detail. Section IV describes the experimental settings and results while Section V concludes the paper.

\section{Related Work}

Our work lies in the intersection between the temporal collaborative filtering (TCF) and the graph-based collaborative filtering literature. In this section, we review these two areas and show the differences between our method and existing ones in each corresponding area. 

\subsection{Temporal Collaborative Filtering \label{sec:tars}}

Existing TCF methods focus on the temporal aspects of only users, of users and items, or of their interactions, e.g., rating scores.
In this section, existing studies are grouped based on how they model users' preferences and items' social perception. Throughout the paper, the opinion that a social environment has of a product will be referred as \textit{item (social) perception}
. 

A number of works consider items' social perception as stationary and model users' preferences as evolving over time~\cite{fusion}. 
For instance, \cite{fusion} uses random walks to independently learn short- and long-term user preferences. 

Nevertheless, most TCF methods in the literature assume that \textit{both} items' perception and users' preferences may evolve over time~\cite{rrn2019,deng2018,Koren2010}. They either (a) model users' and items' latent states by means of temporal MF~\cite{rrn2019,deng2018} or (b) infer rating scores by modeling users' and items' dynamics by means of baseline predictors~\cite{Koren2010}. While~\cite{Koren2010} and \cite{Liu2013} obtain time-stamped latent states by following a MF approach,~\cite{rrn2019} uses Long-Short Term Memory (LSTM) units~\cite{lstm} to learn time-varying functions, thus reconstructing the dynamics of the evolution rather than the latent states.

The previous studies propose methods that learn the evolution of users' preferences and items' perceptions using independent RNNs. Alternatively, other papers define baseline functions which only depend on previous rating scores and incorporate time-decay functions to penalize older ratings~\cite{twcf} or emphasize relevant periods like seasons or week-ends~\cite{rel2}.

Unlike the TCF techniques presented above, which employ baseline functions or MF methodologies, our method models the user-item interactions by means of graph-based techniques, that is, GNNs. Our target is to effectively learn the latent embeddings with these novel neural networks which allow for a more precise study of the interactions and an increase in computational complexity.

\subsection{Graph-based Collaborative Filtering \label{sec:link}}
Link prediction refers to the task of anticipating whether an edge should be created between two nodes. Recommender systems can be formulated as a link prediction problem by representing users, items and their interactions in a bipartite graph. One of the first papers that proposed this methodology was \cite{linkpred} who introduced novel linkage functions as measure for inferring the rating scores in the model. Since then, various methods have followed proposed by \cite{Rianne2017,bnp,convRS,Huang17,Wang2019}. 

Although MF is arguably the most commonly employed approach for recommender systems, recent studies have also leveraged CNNs and GNNs with favorable outcomes. 
Mainly thanks to the development of CNNs and GNNs in recent years, interpreting recommender systems as a link prediction problem has received a boost in attention, leading to performance improvements. Namely, the methods by~\cite{Rianne2017,bnp,Wang2019} outperform non-graph CF papers by incorporating GNNs, whereas \cite{convRS} addresses the problem similarly using CNNs. In effect, these papers leverage the graph structure of the data and employ neural networks in the extraction of the latent embeddings of users and items. Differently, \cite{Huang17} proposed to leverage signal processing  concepts such as as graph frequency and graph filters to address the problem. 

The approaches reviewed in this subsection focus on the static CF problem and ignore the dynamics behind users' opinions and social trends. In this work, we attempt to combine graph-based and time-dependant techniques with the aim of learning user and item latent embeddings and their dynamics.

\section{The Proposed Method}
\begin{figure*}[h]
  \centering
  \includegraphics[scale=0.5]{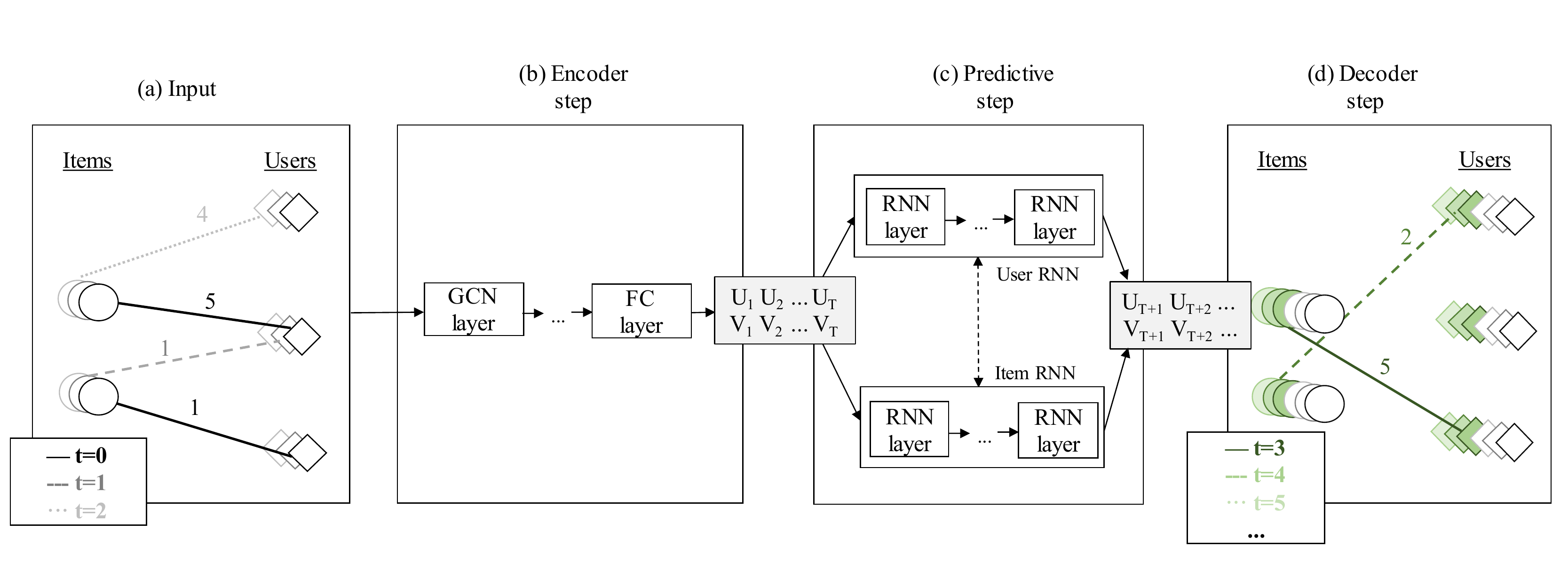} 
  \caption{The proposed time-aware graph-based matrix completion architecture for Recommender Systems (TG-MC). \textbf{(a)} The model receives as input a sparse tensor of ratings R$\in {\rm I\!R}^{N_U \times N_V \times T}$ in three dimensions (users, items and time). The empty elements of R represent unknown rating scores from a user to a specific item. \textbf{(b)} The encoder is formed by combining GNNs and fully-connected layers. Running the encoder step for each time window results in two length-T sequences of latent representations for users and items. \textbf{(c)} The predictive step encompasses an RNN-, LSTM- or GRU-based architecture. Given the two sequences of users and items embeddings, two RNN models, which we refer to as user- and item-RNNs, predict the subsequent embeddings. The two RNNs can share weights or can be completely independent. \textbf{(d)} By matrix multiplication, the decoder step obtains, at the subsequent time instance, the output matrix $\hat{R}_{T+1}$ which contains the predicted ratings scores. 
}
  \label{fig:model}
\end{figure*}
In this work, we follow the TCF approach for recommender systems and consider the time information associated to the user-item interactions. Concretely, we group the observed interactions by the time they occur, resulting in a \textit{sequence of $T$ non-overlapping interaction matrices} $\{R_{1}, \ldots, R_{T} \}$. This is in contrast to the setting in static CF methods, where all the observed interactions are processed at once in a single data matrix. A time step $t$ in our setting is associated to a matrix $R_t$ which contains interactions occurring between $t-\delta/2$ and $t+\delta/2$ with $\delta$ the time window's length. $T$ is hence determined by the total time span of the data (between the first and the last observed interactions) and $\delta$.

To learn from the sequence of interaction matrices, we propose a method, coined Time-aware Graph-based Matrix Completion (TG-MC), which leverages a graph-based CF approach and temporal prediction techniques. The schema of our method is illustrated in Figure~\ref{fig:model}. During training, our method follows a two-stage pipeline. The first stage involves learning latent embeddings representing users and items using the GCMC model~\cite{Rianne2017}. This step is performed \textit{independently} for each interaction matrix $R_t$ with $t = 1\dots T$, resulting in two length-$T$-sequences of latent embeddings for the users and the items. At each time step, a bi-linear decoder is trained together with the GCMC model as in~\cite{Rianne2017}. In the second stage, we fit two RNNs on the sequences of user and item representations to model their temporal trajectories. During testing, these RNNs are used to propagate the user and item representation vectors to a future time instance, from which user-item interaction scores can be computed using the bi-linear decoder trained in the first stage. Next, we present the steps involved in our method in  detail. 
\subsection{Learning Users and Items Latent Embeddings with a Graph-based Model}
\label{sec:user_item_factor_gnn}
\subsubsection{Modeling Users and Items}
Following~\cite{Rianne2017}, we represent the rating matrices in the form of graphs. Concretely, the interaction matrix $R_t$ is represented as a bipartite graph $\mathcal{G}_t$ where a node corresponds to either a user or an item, and an edge encodes a rating a user has given to an item. The aim of this step is to learn the latent representations associated to each user or item node. The latent embedding of the node $i$ at time $t$ is denoted with the vector $z_{i,t} \in {\rm I\!R}^{d \times 1}$, where $d$ is the dimension of the latent representation. 

Independently of the node type (item or user), a latent node embedding $z_{i,t}$ is learned by the following two steps. First, a vector $h_{i,t}$ is computed by accumulating all incoming links of node $i$ (i.e. $j \rightarrow i$, $\forall j$ connected to $i$) in graph $\mathcal{G}_t$
\begin{equation}
    h_{i,t}=\sigma\left[ \text{accum}\left(\sum_{j\in N_{i,1}}  \mu_{j \rightarrow i,1,t} , ... , \sum_{j \in N_{i,R}}   \mu_{j \rightarrow i,R,t} \right) \right],
    \label{eq:acc}
\end{equation}
where $\sigma[\cdot]$ is the activation function and $N_{i,r}$ is the set of nodes with linked interactions to user (or item) $i$ with rating score $r \in \{1,\ldots, R\}$ and $\text{accum}(\cdot)$ is an accumulation function, e.g., concatenation or element-wise addition.
Each incoming link to node $i$ is a user-item interaction and contributes by
\begin{equation}
    \mu_{j \rightarrow i,r,t}=\frac{1}{c_{i,j,t}}W_{r,t} x_{j,t}.
    \label{eq:mpass}
\end{equation}
Here, $W_{r,t} \in {\rm I\!R}^{H \times N_U}$ are learnable weights. $N_U$ (or $N_V$) and $H$ are, respectively, the input size and the hidden dimension size, $x_{j,t}$ is the initial feature vector of item-node $j$ at time $t$, and $c_{i,j,t}$ is the normalization factor.

In the second step, the latent vector $z_{i,t}$ is computed by 
\begin{equation}
    z_{i,t}=\sigma \left( W_t  h_{i,t}\right),
    \label{eq:latrep}
\end{equation}
where each weight matrix $W_t \in {\rm I\!R}^{d \times H}$ is learned from the available training data at time $t$. Depending on the node type (i.e., whether it represents a user or an item), we consider the learned representation $z_{i,t}$ as a user or an item vector, denoted as $u_{i,t}$ and $v_{j,t}$ respectively.

\begin{table*}[t]
\vspace{1cm}
\caption{Average test RMSE and MAE scores for all variations of our method on the Netflix and MovieLens 1M datasets.}
\centering
\begin{tabular}{cc|cc|cc}
\hline
  &                 & \multicolumn{2}{c}{\textbf{Non-Accumulative Representation}} & \multicolumn{2}{|c}{\textbf{Accumulative Representation}} \\
\cline{3-6}
\textbf{Dataset}               & \textbf{Method} & RMSE              & MAE              & RMSE             & MAE            \\
\hline
\hline
\multirow{3}{*}{Netflix} & TG-MC RNN  & $1.008\pm$ 0.085 & 0.921$\pm$0.040&  $1.001\pm0.033$&$0.822\pm0.022$   \\
  & TG-MC GRU       &   0.969 $\pm$ 0.034 &  0.876 $\pm$ 0.037 & $0.952\pm 0.023$  & \textbf{0.767$\pm$0.029}  \\
  & TG-MC LSTM      &     0.974 $\pm$ 0.028 & 0.918 $\pm$ 0.017 & \textbf{0.931$\pm $0.009} & 0.790$\pm $0.020   \\
\hline
\multirow{3}{*}{MovieLens 1M} & TG-MC RNN      &  1.032$\pm$ 0.030 & 0.830$\pm$0.027 & 0.876 $\pm$ 0.023 & 0.783 $\pm$ 0.027  \\
                               & TG-MC GRU     &  1.019$\pm$0.028  &  0.818$\pm$ 0.028 & 0.867 $\pm$  0.013 & 0.697 $\pm$ 0.023 \\
                               & TG-MC LSTM     & 1.015 $\pm$ 0.022 & 0.768$\pm$ 0.026  & \textbf{0.834 $\pm$ 0.011} & \textbf{0.664 $\pm$ 0.028}\\         
\hline
\end{tabular}
\label{table:variations}
\end{table*}

\begin{table*}[t]
\caption{RMSE scores obtained by our method and reference methods on the Netflix and MovieLens 1M datasets.}
\centering
\begin{tabular}{c|cccc}
\hline
\textbf{Method }&\textbf{ Netflix}& \textbf{MovieLens 1M}\\
\hline
\hline
PMF \cite{pmf}& 0.957  & 0.883   \\
I-AutoRec \cite{autorec} & 0.979  & \textbf{0.833} \\
U-AutoRec \cite{autorec} & 0.985 &  0.877  \\
GCMC \cite{Rianne2017} & 1.264&  1.001   \\
TG-MC (ours) & \textbf{0.931} & 0.834\\
\hline
\end{tabular}
\label{table:indep} 
\end{table*}

\subsubsection{Bi-linear Decoder}
Following the GCMC model~\cite{Rianne2017}, for each time window, we compute the user-item interaction scores from the learned user and item embeddings using a bi-linear decoder. In this work, as we consider ratings (e.g., one- to five-score rating) as the interaction type, the prediction of rating values can be treated as a classification problem. We calculate the probability of rating $R_{i,j,t}$ (user $i$ rate item $j$ at time $t$) to have value $r$ according to
\begin{equation}
P (R_{i,j,t}=r)= \frac{\exp (u_{i,t}^T Q_{r,t} v_{j,t})}{\sum_{s \in \mathcal{R}} \exp (u_{i,t}^T Q_{s,t} v_{j,t})} \text{,}
    \label{eq:prob}
\end{equation}
with $u_{i,t}$, $v_{j,t}$ the embeddings of user $i$ and item $j$ at time $t$ and $Q_{r,t} \in {\rm I\!R}^{d \times d}, \forall r=\{1,\ldots,R\}$ the learnable parameters of the bi-linear decoder at time $t$. 

We jointly train the parameters of the bi-linear decoder ($Q_{r,t} \in {\rm I\!R}^{d \times d}, \forall r=\{1,\ldots,R\}$) and those of the graph-based user and item modelling ($W$), separately \textit{at each time window $t$}, by minimizing the negative log-likelihood objective function,
\begin{equation}
    \mathcal{L}_{graph} = - \sum_{i,j \in \Omega_{t}} \log P(R_{i,j,t} = r),
    \label{eq:loss1}
\end{equation}
with $\Omega_{t}$ the set of ratings available for training at time $t$, and $r$ is the ground-truth rating that user $i$ gives to item $j$. 
\subsubsection{Handling Data Sparsity}
CF methods generally suffer from the data sparsity problem, i.e., the number of observed user-item interactions is scant compared to the total possible number of such interactions. 
In our case, this problem aggravates because the number of observed ratings is distributed with respect to their time instance
: if the range of observed interactions is uniformly distributed across time, the density of $R_t$ is on average $\rfrac{1}{T}$ that of $R$. 

To mitigate this problem when building $R_t$, rather than only using the interactions observed within $t$, we also include the interactions accumulated from previous time steps $\{1,\dots,t-1\}$. Following this approach, the density of the rating matrix increases through time, i.e., $|\Omega_{t+1}| > |\Omega_t|, \forall t \in \{1,\dots,T-1\}$, which helps ease the learning of the GCMC model. In Section IV, we will compare the efficiency of this variance (accumulative representation) against the non-accumulative representation.

\subsection{Modeling Users' and Items' Dynamics}
By learning user and item embeddings for each time window~$t$, with $t \in \{1,\dots,T\}$,  we obtain two length-$T$ sequences of embeddings, one for the users and the other for the items. We denote these two sequences, respectively, by $U_t$ and $V_t$ with $t \in \{1, \dots, T\}$. We have $U_t \in  {\rm I\!R}^{N_U \times d}$, $V_t \in  {\rm I\!R}^{N_V \times d}$ where $N_U,N_V$ are the number of users and items.

We employ two RNNs to model the temporal information in these sequences of embeddings, one for the users and one for the items, which we refer to as the user-RNN and item-RNN, respectively. This is motivated by the success of RNN and its variants in learning and predicting sequential data~\cite{sutskever2014}. As we unify the dimensions of the user and item embeddings (both equal to $d$), we can have the user-RNN and item-RNN either sharing weights or operating separately. In Section IV, we will compare the results obtained with these two variants. 
%
%
%

Using RNNs, we predict the embeddings at a time step $T+1$ from the embeddings of the previous $T$ time steps (from $t=1$ to $T$). By re-iterating this process, we can further infer the embeddings at later time steps, e.g., $T+2$ and $T+3$ and so on. Formally, considering the user embeddings, we have:
\begin{equation}
    \widehat{U}_{T+1}= f_{\text{RNN}}(U_1, \ldots, U_T), \\
  \label{eq:predrnn}
\end{equation}
with $\widehat{U}_{T+1}$ the predicted embeddings at time $T+1$ and $f_{\text{RNN}}(\cdot)$ the function implemented by the user-RNN.
The item embeddings at each time window can be predicted in the same way using the item-RNN whose operation is referred as $g_{\text{RNN}}$.


We employ the mean-squared-error (MSE) loss function to train the user- and item-RNNs. Concerning the user-RNN, at time step $t$, the MSE is calculated between the predicted embedding $\widehat{U}_t$ produced by applying the model on the sequence of prior embeddings, $\widehat{U}_1, \dots, \widehat{U}_{t-1}$, and the embedding produced by the GCMC model at time $t$, namely, $U_t$, as illustrated in Eq. (\ref{eq:loss2}). 

%
\begin{equation}
    \mathcal{L}_{\text{user-RNN}} =\dfrac{1}{N}{\sum_{i=1}^{N} \Vert \widehat{U}_i-U_i \Vert_{F}^2},
    \label{eq:loss2}
\end{equation}
with $\Vert . \Vert_{F}$ the Frobenious norm. The loss function used to train the item-RNN is calculated analogously using the item embeddings.
\subsection{Predicting entry values at future time instances}
Using the trained user-RNN and item-RNN models as presented above, we can predict the user and item embeddings at any future time step. Following the same setting, we employ the $T$ bi-linear decoders learnt at time steps $t=\{1,\ldots, T\}$ and a LSTM-based RNN architecture to predict the bi-linear decoder weights at $t=T+1$. The selected loss function is MSE as illustrated in Eq. (\ref{eq:loss2}). With this, we are able to predict the user-item interaction probabilities at $t=T+1$. 
Concretely, consider the time instance $T+1$, the predicted probability for rating $R_{i,j,T+1}$ to have value $r$ is calculated following Eq.~(\ref{eq:prob}), with the embeddings produced by the (user- and item-) RNNs at time $T+1$ and the bi-linear decoder produced by the RNN at time $T+1$. From the computed probabilities, we can produce the continuous-valued predictions according to:
\begin{equation}
    \widehat{R}_{i,j,t}=\sum_{r \in \mathcal{R}} r P (R_{i,j,t}=r)
    \label{eq:expect}
\end{equation}
where $\mathcal{R}=\{1,\ldots, R\}$ is the set of possible rating scores. 
\section{Experiments}
In this section, we present our experimental study to assess the effectiveness of the proposed methods compared to the state of the art. We first describe the experimental settings and then report the obtained results.

\subsection{Experimental Settings}
We employ the Netflix and MovieLens datasets~\cite{Netflix,mldataset} in our experiments,  both widely used in recent literature \cite{rrn2019, Rianne2017, autorec, Alex2011} and with different characteristics regarding number of users, items and distribution of ratings. The Netflix dataset comprises a total of about 100M ratings, $480,189$ users and $17,770$ movies, whereas the MovieLens 1M (ML-1M) dataset has $6,040$ users, $3,900$ items and 1M observed ratings. 

A data point (i.e. rating) of any of the two sets contains a movie and user identifier as well as the rating score and the time of the event. The Netflix timestamp is given as YYYY-MM-DD while the ML-1M is given as seconds since midnight Coordinated Universal Time (UTC) of January 1, 1970. 

For the Netflix dataset, we follow the setup in~\cite{rrn2019} and consider a 6-month subset of the dataset (nearly 5M data points) with ratings occurring between July and December 2005. To respect causality considerations, ratings recorded in the month of December are kept for testing while those recorded between July and November are used for training. All the ratings are integer values in the range $[1-5]$. We split the data according to a time window of $1$ week. As the total time span of the data is $30$ weeks, we obtain $T=30$ interaction matrices, of which the first $26$ ones are used for training (3M ratings) and the rest are reserved for testing (1.8M ratings). This results in a train-test split of 62\%-38\%.

For the ML-1M dataset, we follow~\cite{Alex2011} and split the dataset according to a time window of 3 months. As the total time span of the data ranges from May 2000 to January 2003, we obtain $T=11$ time windows, where the first $9$ are used for training and the remaining are kept for testing,  resulting in a 99\%-1\% train-test split. Like on the Netflix dataset, the ratings are integers in the range $[1-5]$.

We compare the performance of our methods with that of state-of-the-art reference models, including non-temporal CF models such as the PMF~\cite{pmf}, Autorec~\cite{autorec} and GCMC models~\cite{Rianne2017} and temporal CF models~\cite{rrn2019,Koren2010,tcf,ntf,Alex2011,tmf,he2017neural,lfm}. The performance of the models are assessed using two metrics, namely, the root mean squared error (RMSE) and mean absolute error (MAE). For each model, we report the mean results obtained after five different runs employing the test and train sets explained above. 

On the GCMC model \cite{Rianne2017}, the reported results are obtained by running their code on the whole matrix of ratings when $T=1$ with the best hyperparameters reported in their paper. Equivalently, reported results from \cite{rrn2019} on the Netflix dataset are collected from their papers. The temporal MF model from \cite{tmf} was re-implemented and run for the best parameters we found. We followed an equivalent process for \cite{tcf,autorec,he2017neural}.  Results on \cite{ntf,Alex2011,lfm} are taken from their paper since they follow the same experimental settings.

\begin{table*}[t]
\vspace{1cm}
    \caption{Comparison of our method and TCF methods on the Netflix and MovieLens datasets.}
    \begin{subtable}{.47\linewidth}
      \centering
        \caption{Average test RMSE scores for different TCF models on the Netflix Data Set.}
        
\begin{tabular}{ c| c}
\hline
\textbf{Method} & \textbf{RMSE} \\
\hline
\hline
Temporal MF \cite{tmf} & 1.112 \\
RRN \cite{rrn2019}  & 0.944 \\
TimeSVD++ \cite{Koren2010}& 0.962 \\
NCF \cite{he2017neural} & 0.947 \\
LFM \cite{lfm} & 0.936 \\
TG-MC (ours) & \textbf{0.931} \\
\hline
\end{tabular}        

    \end{subtable}%
    \hspace{0.5cm}
    \begin{subtable}{.47\linewidth}
      \centering
        \caption{Average test MAE scores for different TCF models on the Movie-Lens 1M Data Set.}
        
\begin{tabular}{ c| c}
\hline
\textbf{Method}  & \textbf{MAE} \\
\hline
\hline
Temporal MF \cite{tmf} & 0.843\\
RRN \cite{rrn2019} & 0.793\\
AM$^{N=1}$ \cite{Alex2011} & 0.777\\
NTF \cite{ntf} &  0.689 \\
TG-MC (ours) & \textbf{0.664}\\
\hline
\end{tabular}

    \end{subtable} 
    \label{table:all}
\end{table*}


\subsection{Hyperparameters Selection}
We empirically select the hyperparameters for the graph encoder, the user- and the item-RNNs. 

\paragraph{The Graph-based Encoder} For this model, we employ a learning rate of $10^{-2}$, a dropout rate of $0.3$ and rectified linear unit (ReLU) as the activation function after both dense and GNN layers. 

On the Netflix dataset, we performed grid search to select the best combination of the output dimensions of the GNN and dense layers ($H$ and $d$). We compared the results obtained by different combinations on a separate validation set containing $20\%$ of the known ratings in the training set at the last training time step. This procedure results in $H=500$ and $d=75$. For each time step, we train the corresponding graph-based encoder for $1,000$ epochs with a batch size of $100,000$ training ratings. 
On the ML-1M dataset, we perform the same grid search procedure and accordingly set  $H=500$ and $d=50$. For each time step, we train the model for $2,500$ epochs with a batch size of $100,000$ training samples. 


\paragraph{The user- and item-RNN models}
%
%
We experiment with different variants of RNNs to construct the user- and item- recurrent networks, including LSTM, GRU and vanilla RNN models, with one, two or three hidden layers. For each variant, we use the $\tanh$ activation function. As mentioned earlier, the weights between the user- and item- RNNs can be shared. For training, we employ the Adam optimizer with a learning rate of $10^{-2}$, running for $250$ epochs. Throughout our experimental study, we empirically observe that using LSTM models with two hidden layers give the best performance overall, and that sharing weights between the user- and item- recurrent models improves the performance. As such, we use this configuration when comparing our method to the reference models. 
\subsection{Experimental Results}
\paragraph{Non-Accumulative versus Accumulative Representations}Table \ref{table:variations} compares the results obtained by our methods on the Netflix and ML-1M datasets when using the non-accumulative and accumulative data representations. Recall that with the former, only the ratings observed within a time window $t$ are used to learn the user and item embeddings ($U_t$ and  $V_t$ respectively), while with the latter, all ratings between the first and the $t-$th time windows are employed. 

We can observe from the two tables that, given the same experimental conditions and independently of the employed dataset and RNN variant, the accumulative data representation yields better results in terms of RMSE and MAE. Furthermore, it is worth mentioning that more complex units like LSTMs and GRUs provide lower prediction errors than the vanilla RNN.

\paragraph{Comparison against Non-temporal CF models}
In this paper, we argue that (\textit{i}) learning user and item embeddings via GNNs and (\textit{ii}) modeling the temporal dynamics improve the performance of CF models. To confirm both points, we compare the performance of our method with that of both non-temporal and non-graph-based reference CF models. 

Table~\ref{table:indep} shows the RMSE scores obtained by our method and \textit{non-temporal CF reference methods} on the Netflix and ML-1M datasets. As can be seen, our method yields the best performance on the Netflix dataset, followed by the PMF method. On the ML-1M dataset, the I-AutoRec model performs the best while our method has the second best performance. It should be noted that the performance difference between the two is relatively small (i.e., 0.001 point on RMSE on average), and that our method has lower MAE compared to the I-AutoRec model (0.664 versus 0.790, respectively).

Among the reference non-temporal models, the GCMC model~\cite{Rianne2017} employs graph neural network to learn the latent representations. In fact, this model can be seen as a special case of our method where the time span $\delta$ is defined so that all the known rating scores are enclosed in one matrix (i.e., $T=1$). However, directly applying the GCMC model on the TCF setting, where the training and testing sets are split according to the time stamps, results in poor performance on both the Netflix and ML-1M datasets, showing the benefits of modeling the temporal dynamics behind user and item embeddings in TCF. 

\paragraph{Comparison with State-of-the-art TCF models}Table~\ref{table:all} compares the performance of our method and that of reference TCF methods on the Netflix and ML-1M datasets. For a fair comparison, we follow most recent TCF papers \cite{rrn2019,Koren2010,Liu2013} and report RMSE scores for the Netflix dataset, and MAE scores for the ML-1M dataset. It is worth re-calling that whenever applicable, we include the best scores reported in the corresponding papers in our comparison. 
From the table, it is noted that our method yields the best performance on the Netflix dataset. The LFM model achieves the second best RMSE score, followed by the NCF model. On the ML-1M dataset, our method out-performs all reference TCF methods by large margins.

Among the reference TCF models, our method of modeling the temporal trajectories of user and item embeddings is similar that of the Temporal MF model~\cite{tmf}. The key difference is that we employ graph neural network to learn the embeddings at each time step, while the Temporal MF models follow a matrix-factorization approach. The results reported in Table~\ref{table:all} consistently show significant performance improvements obtained by the TG-MC model over the Temporal MF model. This justifies the benefits of using graph neural networks in our method.

\section{Conclusion}

In this paper, we have presented a method for temporal collaborative filtering (TCF), which combines graph neural network (GNN) and recurrent neural network (RNN) models to effectively (\textit{i}) learn the latent user and item representations, and (\textit{ii}) model the trajectories of these representations across time. To deal with the increased data sparsity in the TCF setting, we proposed to train the GNNs using an accumulative data representation technique. 
Our comprehensive experiments on the Netflix and MovieLens 1M datasets justified the benefits of each of the proposed components, namely, the use of RNNs to model the temporal dynamics in the TCF settings, the use of GNNs to capture the latent representations of users and items and the benefits of training the models using accumulative data. The experimental results also showed that our method yielded favourable performance compared to several state-of-the-art TCF models.

\section*{Acknowledgement}
This research was supported by funding from the Flemish Government under the “Onderzoeksprogramma Artificiële Intelligentie (AI) Vlaanderen” programme.

\bibliographystyle{IEEEtran}
\bibliography{IEEEabrv,IEEEexample}

\begin{thebibliography}{10}
\providecommand{\url}[1]{#1}
\csname url@samestyle\endcsname
\providecommand{\newblock}{\relax}
\providecommand{\bibinfo}[2]{#2}
\providecommand{\BIBentrySTDinterwordspacing}{\spaceskip=0pt\relax}
\providecommand{\BIBentryALTinterwordstretchfactor}{4}
\providecommand{\BIBentryALTinterwordspacing}{\spaceskip=\fontdimen2\font plus
\BIBentryALTinterwordstretchfactor\fontdimen3\font minus
  \fontdimen4\font\relax}
\providecommand{\BIBforeignlanguage}[2]{{%
\expandafter\ifx\csname l@#1\endcsname\relax
\typeout{** WARNING: IEEEtran.bst: No hyphenation pattern has been}%
\typeout{** loaded for the language `#1'. Using the pattern for}%
\typeout{** the default language instead.}%
\else
\language=\csname l@#1\endcsname
\fi
#2}}
\providecommand{\BIBdecl}{\relax}
\BIBdecl

\bibitem{cfdos}
Y.~Koren and R.~Bell, ``Advances in collaborative filtering,'' pp. 77--118, 01
  2015.

\bibitem{rrn2019}
C.-Y. Wu, A.~Ahmed, A.~Beutel, A.~Smola, and H.~Jing, ``Recurrent recommender
  networks,'' 02 2017, pp. 495--503.

\bibitem{deng2018}
X.~Ma, P.~Sun, and Y.~Wang, ``Graph regularized nonnegative matrix
  factorization for temporal link prediction in dynamic networks,''
  \emph{Physica A: Statistical Mechanics and its Applications}, vol. 496, pp.
  121 -- 136, 2018.

\bibitem{Koren2010}
Y.~Koren, ``Collaborative filtering with temporal dynamics,'' \emph{Commun.
  ACM}, vol.~53, no.~4, pp. 89--97, Apr. 2010.

\bibitem{song}
Y.~Song, A.~M. Elkahky, and X.~He, ``{Multi-Rate Deep Learning for Temporal
  Recommendation},'' in \emph{Proceedings of the 39th International ACM SIGIR
  Conference on Research and Development in Information Retrieval}, ser. SIGIR
  '16.\hskip 1em plus 0.5em minus 0.4em\relax New York, NY, USA: ACM, 2016, pp.
  909--912.

\bibitem{tcf}
L.~Xiong, X.~Chen, T.-K. Huang, J.~Schneider, and J.~Carbonell, ``Temporal
  collaborative filtering with bayesian probabilistic tensor factorization,''
  12 2010, pp. 211--222.

\bibitem{Liu2013}
N.~N. Liu, L.~He, and M.~Zhao, ``Social temporal collaborative ranking for
  context aware movie recommendation,'' \emph{ACM Trans. Intell. Syst.
  Technol.}, vol.~4, no.~1, Feb. 2013.

\bibitem{timereview}
P.~G. Campos, F.~D{\'i}ez, and I.~Cantador, ``Time-aware recommender systems: a
  comprehensive survey and analysis of existing evaluation protocols,''
  \emph{User Modeling and User-Adapted Interaction}, vol.~24, no.~1, pp.
  67--119, Feb 2014.

\bibitem{fusion}
L.~Xiang, Q.~Yuan, S.~Zhao, L.~Chen, X.~Zhang, Q.~Yang, and J.~Sun, ``{Temporal
  Recommendation on Graphs via Long- and Short-term Preference Fusion},'' in
  \emph{Proceedings of the 16th ACM SIGKDD International Conference on
  Knowledge Discovery and Data Mining}, ser. KDD '10.\hskip 1em plus 0.5em
  minus 0.4em\relax New York, NY, USA: ACM, 2010, pp. 723--732.

\bibitem{ntf}
X.~Wu, B.~Shi, Y.~Dong, C.~Huang, and N.~V. Chawla, ``Neural tensor
  factorization for temporal interaction learning,'' in \emph{Proceedings of
  the Twelfth ACM International Conference on Web Search and Data Mining}, ser.
  WSDM ’19.\hskip 1em plus 0.5em minus 0.4em\relax New York, NY, USA:
  Association for Computing Machinery, 2019, p. 537–545.

\bibitem{Rianne2017}
R.~Van~den Berg, T.~N. Kipf, and M.~Welling, ``{Graph Convolutional Matrix
  Completion},'' \emph{ArXiv}, vol. abs/1706.02263, 2017.

\bibitem{bnp}
S.~Wu, Y.~Tang, Y.~Zhu, L.~Wang, X.~Xie, and T.~Tan, ``Session-based
  recommendation with graph neural networks,'' \emph{Proceedings of the AAAI
  Conference on Artificial Intelligence}, vol.~33, p. 346–353, Jul 2019.

\bibitem{lstm}
S.~Hochreiter and J.~Schmidhuber, ``{Long Short-Term Memory},'' \emph{Neural
  computation}, vol.~9, no.~8, pp. 1735--1780, 1997.

\bibitem{twcf}
P.~Wang, H.~Hou, and X.~Guo, ``{Collaborative Filtering Algorithm Based on User
  Characteristic and Time Weight},'' in \emph{Proceedings of the 2019 8th
  International Conference on Software and Computer Applications}, ser. ICSCA
  '19.\hskip 1em plus 0.5em minus 0.4em\relax New York, NY, USA: ACM, 2019, pp.
  109--114.

\bibitem{rel2}
D.~Lee, S.~E. Park, M.~Kahng, S.~Lee, and S.-g. Lee, \emph{Exploiting
  Contextual Information from Event Logs for Personalized
  Recommendation}.\hskip 1em plus 0.5em minus 0.4em\relax Berlin, Heidelberg:
  Springer Berlin Heidelberg, 2010, pp. 121--139.

\bibitem{linkpred}
H.~{Chen}, X.~{Li}, and Z.~{Huang}, ``{Link Prediction Approach to
  Collaborative Filtering},'' in \emph{Proceedings of the 5th ACM/IEEE-CS Joint
  Conference on Digital Libraries (JCDL '05)}, June 2005, pp. 141--142.

\bibitem{convRS}
D.~Kim, C.~Park, J.~Oh, S.~Lee, and H.~Yu, ``Convolutional matrix factorization
  for document context-aware recommendation,'' in \emph{Proceedings of the 10th
  ACM Conference on Recommender Systems}, ser. RecSys ’16.\hskip 1em plus
  0.5em minus 0.4em\relax New York, NY, USA: Association for Computing
  Machinery, 2016, p. 233–240.

\bibitem{Huang17}
W.~{Huang}, A.~G. {Marques}, and A.~{Ribeiro}, ``Collaborative filtering via
  graph signal processing,'' in \emph{2017 25th European Signal Processing
  Conference (EUSIPCO)}, ser. WSDM ’19, 2017, pp. 1094--1098.

\bibitem{Wang2019}
X.~Wang, X.~He, M.~Wang, F.~Feng, and T.-S. Chua, ``Neural graph collaborative
  filtering,'' \emph{Proceedings of the 42nd International ACM SIGIR Conference
  on Research and Development in Information Retrieval - SIGIR’19}, 2019.

\bibitem{pmf}
R.~Salakhutdinov and A.~Mnih, ``{Probabilistic Matrix Factorization},'' in
  \emph{Proceedings of the 20th International Conference on Neural Information
  Processing Systems}, ser. NIPS'07.\hskip 1em plus 0.5em minus 0.4em\relax
  USA: Curran Associates Inc., 2007, pp. 1257--1264.

\bibitem{autorec}
S.~Sedhain, A.~K. Menon, S.~Sanner, and L.~Xie, ``Autorec: Autoencoders meet
  collaborative filtering,'' in \emph{Proceedings of the 24th International
  Conference on World Wide Web}, ser. WWW '15 Companion.\hskip 1em plus 0.5em
  minus 0.4em\relax New York, NY, USA: ACM, 2015, pp. 111--112.

\bibitem{sutskever2014}
I.~Sutskever, O.~Vinyals, and Q.~V. Le, ``Sequence to sequence learning with
  neural networks,'' 2014.

\bibitem{Netflix}
``Netflix {P}rize {D}ata {S}et,''
  \url{http://archive.ics.uci.edu/ml/datasets/Netflix+Prize}, 2009.

\bibitem{mldataset}
F.~M. Harper and J.~A. Konstan, ``The movielens datasets: History and
  context,'' \emph{ACM Trans. Interact. Intell. Syst.}, vol.~5, no.~4, Dec.
  2015.

\bibitem{Alex2011}
A.~Karatzoglou, ``Collaborative temporal order modeling,'' 10 2011, pp.
  313--316.

\bibitem{tmf}
Y.-Y. Lo, W.~Liao, and C.-S. Chang, ``Temporal matrix factorization for
  tracking concept drift in individual user preferences,'' \emph{IEEE
  Transactions on Computational Social Systems}, vol.~PP, 10 2017.

\bibitem{he2017neural}
X.~He, L.~Liao, H.~Zhang, L.~Nie, X.~Hu, and T.-S. Chua, ``Neural collaborative
  filtering,'' 2017.

\bibitem{lfm}
X.~Shi, X.~Luo, M.~S. Shang, and L.~Gu, ``Long-term performance of
  collaborative filtering based recommenders in temporally evolving systems,''
  \emph{Neurocomputing}, vol. 267, 06 2017.

\end{thebibliography}
\newpage

\end{document}